\documentclass[sigconf]{acmart} 

\usepackage{subcaption}
\usepackage{multirow}
\usepackage{hyperref}
\usepackage{colortbl}
\usepackage{soul}
\usepackage{xr}

\usepackage[]{algpseudocode}
\usepackage{algorithm}
\algnewcommand\algorithmicforeach{\textbf{for each}}
\algdef{S}[FOR]{ForEach}[1]{\algorithmicforeach\ #1\ \algorithmicdo}

\AtBeginDocument{%
  \providecommand\BibTeX{{%
    \normalfont B\kern-0.5em{\scshape i\kern-0.25em b}\kern-0.8em\TeX}}}

\copyrightyear{2023}
\acmYear{2023}
\setcopyright{rightsretained}
\acmConference[GECCO '23 Companion]{Genetic and Evolutionary Computation Conference Companion}{July 15--19, 2023}{Lisbon, Portugal}
\acmBooktitle{Genetic and Evolutionary Computation Conference Companion (GECCO '23 Companion), July 15--19, 2023, Lisbon, Portugal}
\acmDOI{10.1145/3583133.3590655}
\acmISBN{979-8-4007-0120-7/23/07}

\begin{document}

% \title{Auto-Dynamic Particle Swarm Optimization}
\title{AbCD: A Component-wise Adjustable Framework for Dynamic Optimization Problems}

\author{Alexandre Mascarenhas}
%\email{mascarenhasav@gmail.com}
\affiliation{%
  \institution{University of Tsukuba}
  \city{Tsukuba}
  \state{Ibaraki}
  \country{Japan}
}

\author{Yuri Lavinas}
%\email{lavinas.yuri.xp@alumni.tsukuba.ac.jp}
\orcid{0000-0003-2712-5340}
\affiliation{%
  \institution{University of Tsukuba}
  \city{Tsukuba}
  \state{Ibaraki}
  \country{Japan}
}

\author{Claus Aranha}
%\email{caranha@cs.tsukuba.ac.jp}
\affiliation{%
  \institution{University of Tsukuba}
  \city{Tsukuba}
  \state{Ibaraki}
  \country{Japan}
}

\renewcommand{\shortauthors}{A.Mascarenhas, C.Aranha, Y.Lavinas}

\begin{abstract}

Dynamic Optimization Problems (DOPs) are characterized by changes in the fitness landscape that can occur at any time and are common in real world applications. The main issues to be considered include detecting the change in the fitness landscape and reacting in accord. Over the years, several evolutionary algorithms have been proposed to take into account this characteristic during the optimization process. However, the number of available tools or open source codebases for these approaches is limited, making reproducibility and extensive experimentation difficult. To solve this, we developed a component-oriented framework for DOPs called Adjustable Components for Dynamic Problems (AbCD), inspired by similar works in the Multiobjective static domain. Using this framework, we investigate components that were proposed in several popular DOP algorithms. Our experiments show that the performance of these components depends on the problem and the selected components used in a configuration, which differs from the results reported in the literature. Using irace, we demonstrate how this framework can automatically generate DOP algorithm configurations that take into account the characteristics of the problem to be solved. Our results highlight existing problems in the DOP field that need to be addressed in the future development of algorithms and components.

\end{abstract}





%%  \includegraphics[width=\textwidth]{sampleteaser}
%%  \caption{Seattle Mariners at Spring Training, 2010.}
%%  \Description{Enjoying the baseball game from the third-base
%%  seats. Ichiro Suzuki preparing to bat.}

\maketitle
%component-based perspective
\section{Introduction}

Dynamic Optimization Problems (DOPs) are problems where the fitness landscape changes over time. Because of this, a solution can have multiple fitness values as the search progresses. DOPs include vehicle routing, which changes depending on traffic conditions, 
% CLAUS: changed robot control to the problem that Alexandre wants to work with.
and team scheduling, which changes as task requests arrive. Many real-world problems have dynamic characteristics~\cite{survey-phd-2011}, making the study of DOPs crucial.

Different from static problems, Evolutionary Algorithms (EAs) applied to DOPs have to adapt to changes in the fitness functions in the middle of the optimization process. So the EA needs to detect the change to the fitness function and find the new optimum quickly. 

Several EAs and variations have been proposed in the DOP literature in recent years~\cite{algo-realoc-2009, CPSO-2010, mQSO-2006, FTMPSO-2013, Survey_dynamicEAs-A}. However, there is a lack of analysis of individual algorithm components and their interactions. One of the main reasons for this is the lack of source codes~\cite{Survey_dynamicEAs-B}. This causes problems with reproducibility and diminishes our progress and ability to build upon previous knowledge.

In this work, we propose a component-wise framework for DOPs, which we call the \emph{AbCD}: an \underline{A}djusta\underline{b}le \underline{C}omponent framework for \underline{D}ynamic Problems. Our goals with this framework are three-fold: 1) to make available an open-source tool that facilitates the reproducible analysis of several DOP algorithms and experiments; 2) to examine individual components from the literature, with their individual impact and interaction with other components and problem classes; and 3) to design new DOP algorithm configuration, both manually and by automatically searching from the available components and their parameters.

We validate the proposed framework by applying it to the Moving Peaks DOP Benchmark (MPB)~\cite{MPB-1999}. We design two algorithm configurations, one manually by the analysis of individual components, and another automatically by using Iterated Racing~\cite{LOPEZIBANEZ201643}. We compare both configurations, as well as some algorithms from the literature, using standard DOP metrics. The individual analysis of the components also shows interesting insights about existing DOP EAs, and points to promising directions of how to improve the state of the art.
\section{Related Works}\label{related-work}

%Much work on DOPs has been published in the last 20 years.
%, mainly due to the great relevance of this field to real-world problems. 
Many algorithms have been proposed for DOPs, most of them based on Particle Swarm Optimization (PSO), such as FTmPSO~\cite{FTMPSO-2013}, RPSO~\cite{RPSO}, CPSO~\cite{CPSO-2010}, and mQSO~\cite{mQSO-2006}. Their approach is to add dynamic components to PSO that improve its performance in dynamic environments. In particular, mQSO was one of the first to use dynamic components to deal with fitness landscape changes over the optimization time, while the other algorithms were strongly influenced by its ideas. 
However, in spite of the large number of proposed algorithms, there was little focus on creating and sharing tools to enable their use and study, a problem that was well stated in 2013~\cite{related1-2013}, and still placed as one of the main issues in DOPs by 2021~\cite{Survey_dynamicEAs-B}. 

A very small number of works use a component-oriented approach to DOPs. For example, Haluk et al. apply Hyper-heuristics to the Memory/Search algorithm to improve characteristics that are important for DOPs, such as population diversity~\cite{example2-2014}. Danial et al. defines three strategies (Cooperative Evolutionary, Tracking multiple moving optima, and Resource allocation) that are applied to several optimizers such as CMA-ES, jDE, DynDE and PSO~\cite{example3-2019}.
Even in these cases, the problems due to the absence of source code remain, such as reproducibility and the difficulty of analyzing components individually as well as their interaction effects, although \cite{example3-2019} discuss in detail the differences between using or not certain strategies on the different algorithms.

%There are few works aimed at allowing an analysis of the composition of algorithms for DOPs, some were proposed and tested, such as \cite{example1-2019} shows a framework where six different strategies are used in EAs (PSO and DE) in order to obtain better results in different types of DOPs. \cite{example2-2014} presents a framework for DOPs where Hyper-heuristic techniques are applied, which will be heuristics capable of adapting during the problem resolution, in the memory/search (MS) algorithm, in order to make it have desirable characteristics in DOPs, such as population diversity and subpopulations.
%In \cite{example3-2019} a deeper approach to the analysis of the components that compose the algorithm is made. Where three approaches, namely Cooperative Evolutionary, Tracking multiple moving optima, and Resource Allocation, are used in different optimizers (CMAES, jDE, DynDE and PSO) to analyze their impact on the performance of the algorithm.

% However, even in these, some of the main problems still remain, for example, the difficulty of reproducibility due to the absence of the source code and the possibility of analyzing the components individually as well as the compatibility between different components with the optimizer, with a caveat for the last which doesn't quite analyze compatibility but provides detailed information on the difference between using or not certain approaches.
\section{Performance measures}\label{prelim}

A \emph{environment change} happens when the parameters of the objective function change in a dynamic problem. Let $NE \in \mathbb{N}$ be the total number of evaluations in one run of an algorithm, an environment change can occur at any evaluation $n \in [1,NE] \subset \mathbb{N}$. Let $CE \in \mathbb{N}$ be the total number of changes that can occur ($CE < NE$), each $e \in [1,CE] \subset \mathbb{N}$ represents an environment in the run. $N(e) \in \mathbb{N}$ is the total number of fitness evaluations on the $e^{th}$ environment.

\emph{Subpopulations} are often used in algorithms for dynamic optimization. Let $pop \in \mathbb{N}$ the total number of individuals in the population, and $N_{subpops} \in \mathbb{N}$ the number of subpopulations, and therefore, $N_{subpops} < pop$, and each $subpop \in [1,N_{subpops}] \subset \mathbb{N}$ represents a subpopulation. 

So, the term $I_{i,subpop}(e, n)$ is the individual $i$ of the subpopulation $subpop$ at an environment $e$ at the evaluation $n$; The term $I_{best_{i,subpop}}(e, n)$ is used to refer to the position with the highest fitness an individual $I_{i,subpop}$ has had at the environment $e$ up to evaluation $n$; The term $S_{best_{subpop}}(e, n)$ is the best individual of the subpopulation $subpop$ at the environment $e$ up to evaluation $n$; and the term $P_{best}(e, n)$ is used to refer to the best individual among the entire population $pop$ at the environment $e$ up to evaluation $n$.

The metrics implemented in the AbCD Framework is the most frequently used in the field of dynamic optimization, the Offline error ($E_o$) \cite{offline-error-2002}, which is calculated as the average of the best results found up to a given number of evaluations. Its Equation is as follows:

\begin{equation}
    E_{o} = \frac{1}{NE^{'}}{\sum_{n=1}^{NE^{'}} |f(G_{Op}(n)) - f(P_{best}(n))|}
\end{equation}

where $f(G_{Op}(n))$ is the fitness value of the global optimum in the current environment at the $nth$ fitness evaluation, $f(P_{best}(n))$ is the fitness value of the best individual at the $nth$ fitness evaluation and $NE^{'}$ is the number of fitness evaluations so far. During the execution of the algorithm $NE^{'}$ is less than $NE$, and at the end of the execution, we have $NE^{'} = NE$.
\section{A\MakeLowercase{b}CD: Adjustable Components for Dynamic Problems}\label{abcd}

In this work, we create the \underline{A}djusta\underline{b}le \underline{C}omponent framework for \underline{D}ynamic \underline{P}roblems based on commonly used components of dynamic evolutionary algorithms. We designed AbCD  following the component-wise framework, similar to the protocols used in the multiobjective domain~\cite{bezerra2016automatic,moeadr_paper,lavinas_gecco2022}. 

The framework components are categorized according to how they act on individuals in the population. The categories are Local (L), when it involves only one subpopulation or Global (G) when it involves more than one subpopulation. The list of all components in AbCD is shown in Table~\ref{dynamical-components}. Each component can be classified given their functions: optimizer; change detection; convergence detection; diversity control; and population division and management~\cite{Survey_dynamicEAs-A}.

\begin{table}[htbp]
\centering
	%\footnotesize
    \scriptsize
	\caption{Dynamic components in the AbCD framework}
	\label{dynamical-components}
    \begin{tabular}{l|l|l}
        % \hline
        
        \rowcolor[gray]{.85}Component Level    & Component              & Parameter(s)  \\ 
        \multirow{2}{*}{Global component}      & Multipopulation           & $N_{subpops}$    \\
                                    & Exclusion                 & $r_{excl}$    \\ \hline
        % \multirow{3}{*}{Local component}       & Reevaluation $S_{best}$   & $N/A$           \\ 
        \multirow{3}{*}{Local component}       & Reevaluation $S_{best}$   & None           \\ 
                                               & Anti-convergence          & $r_{conv}$ \\ 
                                               & Local Search              & $r_{ls}$, $etry$ \\ \hline    
        \multirow{3}{*}{Optimizer}             & ES                        & $r_{cloud}$   \\
                                               & PSO                       & $\phi_1$, $\phi_2$\\
                                               & Hybrid(PSO+ES)            & $\phi_1$, $\phi_2$, $\%ES_{ind}$, $r_{cloud}$ \\ %\hline    
        \end{tabular}
        % \vspace{-1.5em}
\end{table}

\section{Experiments and Results}\label{experiments}

To evaluate the components as well as test the generated algorithms, the Moving Peaks Benchmark (MPB) was used, as it is available from the Python library DEAP~\cite{DEAP_JMLR2012}. This benchmark has several parameters configurations, for example, the number of peaks, frequency of change, and severity of the change, among others. Among the predefined scenarios,we set MPB to the scenario 2, as it is frequently used~\cite{survey-phd-2011, mQSO-2006, CPSO-2010, FTMPSO-2013, GMPB-2022}.

For all the experiments in this section, the metrics presented are calculated over the results of $50$ runs, each one with $500000$ fitness evaluation. Only the change detection component is always enabled during all experiments, as it is necessary in any DOP. The metrics used to evaluate the configurations are the offline error ($E_{o}$).

\subsection{Baseline}

As a baseline, we configure PSO and ES in the AbCD framework. The PSO parameters values used are $\chi = 0.729$, $\phi_{1}=2.05$, and $\phi_{2}=2.05$, these are commonly used in the literature, and more details about these values can be seen in~\cite{pso-parameters-1, pso-parameters-2}.

Now, for the ES, and hybrid variations, we set the parameter value $r_{cloud}$ experimentally. We looked at the performance of $AbCD_{ES25}$, $AbCD_{ES50}$, $AbCD_{ES75}$ and $ES$ in terms of variable values of $r_{cloud}$ in the MPB set.
\subsection{Manual Design}

To design the Manual Designed AbCD configuration, $AbCD_{man}$, we use the settings of MPB shown in Table \ref{mpb-settings-manual}, and the range of values for each component are shown in Table \ref{manual_parameter_table}. Each component was evaluated individually, as the only addition to the main base algorithm, for each optimizers. The number of individuals in a population is fixed in $pop=100$, a commonly used value from the literature. The value $etry=20$ for the Local search component was chosen based on the experiments in \cite{FTMPSO-2013}. 
\begin{table}[htbp]

\centering
	%\footnotesize
    \scriptsize
	\caption{MPB settings for baseline and manual design}
	\label{mpb-settings-manual}
    \begin{tabular}{l|l}
        % \hline
        \rowcolor[gray]{.85}Parameter                  & Setting           \\ 
        \multirow{1}{*}{Peak function}                 & $cone()$\cite{DEAP_JMLR2012}    \\ \hline     
        \multirow{1}{*}{Number of peaks}               & $10$    \\ \hline    
        \multirow{1}{*}{Number of dimensions}          & $10$ \\ \hline    
        \multirow{1}{*}{Peak heights}                  & $\in [30, 70]$    \\ \hline    
        \multirow{1}{*}{Peak widths}                   & $\in [1, 12]$    \\ \hline  
        \multirow{1}{*}{Change frequency}              & Every $5000$  evaluations    \\ \hline    
        \multirow{1}{*}{Change severity $s$}                & $1$                                  \\ \hline    
        \multirow{1}{*}{Correlation coefficient $\lambda$}  & $0$ \\ 
        \end{tabular}
        % \vspace{-1.5em}
\end{table}

\begin{table}[htbp]

\centering
	%\footnotesize
    \scriptsize
	\caption{Components search space for manual design}
	\label{manual_parameter_table}
    \begin{tabular}{l|l}
        % \hline
        
        \rowcolor[gray]{.85}Components & Domain            \\ 
        \multirow{1}{*}{Optimizer}      & $PSO, AbCD_{ES25}, _{ES50}, _{ES75}, ES$ \\ \hline
        \multirow{1}{*}{Multipopulation}    & $N_{subpops}=1, 5, 10, 25, 50, 100$    \\  \hline  
        \multirow{1}{*}{Exclusion}          & $r_{excl} \in [0, 50]$    \\    \hline    
        \multirow{1}{*}{Anti-convergence}   & $r_{conv} \in [0, 50]$    \\      \hline    
        \multirow{1}{*}{Local Search}       & $r_{ls} \in [0, 50]$ and $ etry=20$    \\  %\hline  
        \end{tabular}
        % \vspace{-1.5em}
\end{table}

We first analyze the multipopulation component. Increasing the sub-population size deteriorates the performance of the AbCD variants and that for the pair instances and configurations tested, having $N_{subpops} = 1$, or just a single population, leads to the best results in terms of offline error.

That said, most of the dynamic components (with the exception of Local search) depend on the configurations using multipopulations. Therefore, for the setting of the parameters of such components, we follow the suggested value of the number of subpopulations being equal to the number of peaks in the problem, $N_{subpops} = 10$,~\cite{mQSO-2006} .

Interestingly, we can only observe small differences for low values of radii while higher values lead to higher error values.  Given the limited positive impact of the dynamic components, we select the $AbCD_{man}$ as the $AbCD_{ES75}$ without dynamic components.

\subsection{Automatic Design}

To design the automatic AbCD configuration ($AbCD_{auto}$) we use different settings of MPB for training and testing, as shown in Table \ref{mpb-settings-automatic}. The range of values for each component are shown in Table \ref{automated_parameter_table}.

\begin{table}[htbp]

\centering
	%\footnotesize
    \scriptsize
	\caption{MPB settings for automatic \textbf{training} and \textbf{testing}}
	\label{mpb-settings-automatic}
    \begin{tabular}{c|c|c}
        \rowcolor[gray]{.85}Parameter                  & training & testing            \\ 
        \multirow{1}{*}{Peak function}                 & $cone()$ & $cone()$ \\ \hline    
                \multirow{1}{*}{Peaks}                 & $8, 10$  & $9, 11$  \\ \hline    
        \multirow{1}{*}{Dimensions}                    & $8, 10$ & $7,9$ \\ \hline    
        \multirow{1}{*}{Peak heights}                  & $\in [30, 70]$  & $\in [30, 70]$  \\ \hline  
        \multirow{1}{*}{Peak widths}                   & $\in [1, 12]$& $\in [1, 12]$    \\ \hline  
        \multirow{1}{*}{Change frequency}              & Every $5000$  evals  & Every $5000$  evals   \\ \hline    
        \multirow{1}{*}{Change severity $s$}           & $1, 2$          & $1.5, 2.5$                                  \\ \hline    
        \multirow{1}{*}{Correlation $\lambda$}         & $0$ & $0$ \\ 
        \end{tabular}
\end{table}

% \begin{table}[htbp]

% \centering
% 	\small
% 	\caption{MPB settings for \textbf{manual} and automatic \textbf{training} and \textbf{testing}}
% 	\label{mpb-settings}
%     \begin{tabular}{c||c||c|c}
%         \rowcolor[gray]{.85}Parameter                  & manual  & training & testing            \\ 
%         \multirow{1}{*}{Peak function}                 & $cone()$\cite{DEAP_JMLR2012}   & $cone()$ & $cone()$ \\ \hline    
%         \multirow{1}{*}{Peaks}               & $10$ & $8, 10$  & $9, 11$  \\ \hline    
%         \multirow{1}{*}{Dimensions}          & $10$ & $8, 10$ & $7,9$ \\ \hline    
%         \multirow{1}{*}{Peak heights}                  & $\in [30, 70]$  & $\in [30, 70]$  & $\in [30, 70]$  \\ \hline  
%         \multirow{1}{*}{Peak widths}                   & $\in [1, 12]$& $\in [1, 12]$& $\in [1, 12]$    \\ \hline  
%         \multirow{1}{*}{Change frequency}              & Every $5000$  evals & same  & Every $5000$  evals   \\ \hline    
%         \multirow{1}{*}{Change severity $s$}                & $1$ & $1, 2$          & $1.5, 2.5$                                  \\ \hline    
%         \multirow{1}{*}{Correlation $\lambda$}  & $0$ & $0$ & $0$ \\ 
%         \end{tabular}
% \end{table}

\begin{table}[htbp]
\centering
	%\footnotesize
    \scriptsize
	\caption{Components search space for automatic design}
	\label{automated_parameter_table}
    \begin{tabular}{l|l}
    % {p{1.3cm}|p{1.4cm}}
        \rowcolor[gray]{.85}Components                 & Domain            \\ 
        \multirow{1}{*}{Population size}               & $100$, $150$, $200$, $300$ \\\hline      
        \multirow{5}{*}{Optimizer}                     &  $PSO$ $\rightarrow \phi_1 \in [0, 2.50], \phi_2 \in [0, 2.50]$ \\
                                                       & $ES$ $\rightarrow r_{cloud} \in [0, 5]$ \\
                                                        \cline{2-2} 
                                                       & \begin{tabular}{l}
                                                           $AbCD_{ES{\%}}$ $\rightarrow$ 
                                                           $\begin{cases}
                                                                \% = 25, 50, 75 \\ 
                                                                \phi_1 \in [0, 2.50], \phi_2 \in [0, 2.50] \\ 
                                                                r_{cloud} \in [0,5]
                                                            \end{cases}$
                                                       \end{tabular} \\ \hline
        \multirow{2}{*}{Multipopulation}    & True, $N_{subpops}=10, 25, 50, 100$    \\    
                                            & False, not used \\ \hline    
        \multirow{2}{*}{Exclusion}          & True, $r_{excl} \in [0, 80]$    \\    
                                            & False, not used \\ \hline    
        \multirow{2}{*}{Anti-convergence}   & True, $r_{conv} \in [0, 80]$    \\      
                                            & False, not used \\ \hline    
        \multirow{2}{*}{Local Search}       & True, $r_{ls} \in [0, 80], etry \in [1,50]$    \\      
                                            & False, not used \\ %\hline    
        
        \end{tabular}
\end{table}

We were surprised by the $AbCD_{auto}$ configuration in terms of the number of sub-populations and the Local search values. Having a large number of sub-population seems to benefit $AbCD_{auto}$ by reducing the interaction among the individuals in the population, probably allowing $AbCD_{auto}$  to explore more peaks. Also, this configuration has Local search with a high $r_{ls}$ value and a small number of iterations ($etry = 2$), which suggests that Local search is used to explore a wider area near the best individual.  

When we compare the results on $AbCD_{auto}$ with $AbCD_{man}$ we see some consensus for most of the components. For example, both configurations use 75\% ES individuals (with 25\% being PSO individuals). Also, they don't use the dynamic components exclusion and anti-convergence. The PSO and ES parameters are slightly different. Moreover, the most divergent choices of components are the number of sub-populations, in different extremes, and the presence of local search in $AbCD_{auto}$. Although we can see a difference in the use of the local search component, irace selected a small number of tries, which suggests that the impact of this component is likely to be small. On the other hand, the reasons behind the contrary number of sub-populations needs more work.

\subsection{Results}

Here we compare PSO, ES, mQSO, $AbCD_{man}$, and $AbCD_{auto}$ using the offline error ($E_o$). We evaluate them in five instances, generated under the MPB scenario 2 setting. We refer to them first by their dimension, then by the number of peaks, and finally by the severity. For example, the 5D-10P-1s instance with 5 dimensions, 10 peaks, and severity 1. We show the results in Table~\ref{aee}.
\begin{table}[htbp]
\centering
	\footnotesize
	\caption{$E_{o}$ of the different configurations}
	\label{aee}
    \begin{tabular}{c|c|c|c}
        \rowcolor[gray]{.85}Algorithm & 5D-10P & 8D-10P & 10D-10P \\ \hline
        $PSO$ & 
            $E_{o} = 15.12(1.62)$ & 
            $E_{o} = 13.01(2.43)$ & 
            $E_{o} = 19.42(0.03)$ \\
            \hline
        $ES$ & 
            $E_{o} = 14.63(1.71)$ &  
            $E_{o} = 13.17(2.71)$ & 
            $E_{o} = 19.04(0.12)$ \\
            \hline
        $mQSO$ & 
            $E_{o} = 7.97(1.21)$ &
            $E_{o} = 13.64(0.66)$ & 
            $E_{o} = 23.36(0.89)$ \\
            \hline
        $AbCD_{man}$ & 
            $E_{o} = 14.78(1.71)$ & 
            $E_{o} = 12.80(2.53)$ & 
            \hl{$E_{o} = 18.72(0.03)$} \\
            \hline
        $AbCD_{auto}$ & 
            \hl{$E_{o} = 6.37(0.71)$} & 
            \hl{$E_{o} = 11.10(1.23)$} & 
            $E_{o} = 19.90(2.49)$ \\                  
        \end{tabular}
\end{table}

For lower dimensions, 5D-10P-1s instance, we can see that the performance of the configurations can be divided into two groups, those with a good performance value, with mQSO and $AbCD_{auto}$, and the second group with the other configurations. The performance of both mQSO and $AbCD_{auto}$ deteriorates as the number of dimensions increases, although $AbCD_{auto}$ keeps a good overall performance. 

This good performance of $AbCD_{auto}$ comes without a surprise since irace searched for a configuration that would perform well in most of the instances. Interestingly, $AbCD_{auto}$ was designed for instances with high dimensions, but it could extrapolate its good performance to an easier problem. We ask ourselves if the same would be true for more challenging instances.

A more interesting result is that dimensionality has a big impact on the performance of the configurations. For example, mQSO is among the best in the problem with the lowest dimension and the worse as the number of dimensions increases. This goes in agreement with the  results of the manual and automatic design, which found that using the main components of mQSO, exclusion, and anti-convergence, lead to a reduction in the performance of the configurations. We believe one of the reasons for this loss in performance in high dimensions could be given that these two components use the Euclidean distance and a more suited metric for distances in many dimensions should be considered in future works.

Another remarkable result is that a recombination of commonly used evolutionary algorithms in static problems can lead to good dynamic configurations, as shown by the increments in performance by the configurations $AbCD_{man}$ and $AbCD_{auto}$. This suggests that incorporating even more effective evolutionary algorithms into the AbCD framework would benefit the dynamic community.

\section{Conclusion}\label{discussion}

% \begin{itemize}
%     \item dynamic components didn't perform well here - do we need them? or do we need more work for specific components? - ok
%     \item peaks vs dimensions - add this to the results
%     \item dimension is causing a problem - ok
%     \item high dimensionality vs dynamical features - ????
%     \item Hybrid - ES (quantum from lit) and PSO, both in manual and automatic - ok
%     \item Local search in automatic design not in manual design -1 pop vs 25 sub-pops - 8 or 10 peaks - ok
%     \item anti-convergence was in the elite, not the best - ok
% \end{itemize}

The aim of this work was to introduce a new component-wise based on commonly used operators in DOPs, the \underline{A}djusta\underline{b}le \underline{C}omponent framework for \underline{D}ynamic \underline{P}roblems. The main goals were to make it available as a framework for extensive experimentation of dynamic EAs and the analysis of new dynamic components and to facilitate the design of ones as well as a tool to simplify the manual and automatic design of algorithms.  All the code and experimental scripts as well the most current version of the framework is available and can be found in GitHub \href{}{https://github.com/mascarenhasav/AbCD}\footnote{The source code for the framework, as well as experimental data used in this paper, is available in the supplementary materials. It will also be on an online repository by the final submission}.

 This analysis allowed us to verify that, contrary to our expectations, the dynamic components were less influential in leading to increments in the performance of the algorithm configurations studied here. It seems that the effect can only be observed in problems with low dimensionality. We understand that this suggests that more work should be done to improve their influence in more dimensions. We also found that the dimensionality of the problem seems to have a higher impact on the search ability of the EAs studied in this work.
 
 We conducted an analysis on the choices of components in the $AbCD_{man}$ and the $AbCD_{auto}$, and we observed that there is an agreement in the choice of the percentage of the number of individuals that are optimized by ES or PSO and in the choice of the dynamic components, not selected in both configurations. However, it seems that there is some interaction among some components since the choice of Local search and the number of sub-population is clearly different between these two configurations. More focus should be directed towards clarifying these interactions.
 
%%%%%%%%%%%%%%%%%%%%%%%%%%%%%%%%%%%%%%%%%%%%%%%%%%%%%%%%%%%%
 
 Then, we studied PSO, ES, mQSO, $AbCD_{man}$, and $AbCD_{auto}$. The first two algorithms are commonly used metaheuristics, mQSO is a representative of EAs for dynamic problems, and the $AbDC_{man}$ and $AbCD_{auto}$ are algorithm configurations generated using the components available in our framework. The results have shown that their relative efficiency depends on the problem's difficulty: mQSO and $AbCD_{auto}$ are the best in lower dimensional problems while PSO, ES, and $AbCD_{man}$ achieve a higher performance as the number of dimensions increase. These results show the power of the AbCD framework, since we can develop established and new EAs and study their performance in dynamic problems.

 %%%%%%%%%%%%%%%%%%%%%%%%%%%%%%%%%%%%%%%%%%%%%%%%%%%%%%%%%%%
%This study strengthens the view using a component-based perspective
%can lead us to a better understanding of the reasons behind variations
%in the performance of different algorithms. One limitation of
%our work is that the number of components and optimizers in the AbCD framework is limited and future works should focus on finding good components candidates, mainly from effective dynamic algorithms from the literature, as well as introducing new components with insights from our results.

 %%%%%%%%%%%%%%%%%%%%%%%%%%%%%%%%%%%%%%%%%%%%%%%%%%%%%%%%%%%
 % Overall, this study strengthens the view using a component-based perspective can lead us to a better understanding of the reasons behind variations in the performance of different algorithms.

% \section{Acknowledgments}

% \section{Appendices}


%\newpage
%\clearpage
\bibliographystyle{ACM-Reference-Format}
\bibliography{references}


\begin{thebibliography}{21}

%%% the Web address.

\ifx \showCODEN    \undefined \def \showCODEN     #1{\unskip}     \fi
\ifx \showDOI      \undefined \def \showDOI       #1{#1}\fi
\ifx \showISBNx    \undefined \def \showISBNx     #1{\unskip}     \fi
\ifx \showISBNxiii \undefined \def \showISBNxiii  #1{\unskip}     \fi
\ifx \showISSN     \undefined \def \showISSN      #1{\unskip}     \fi
\ifx \showLCCN     \undefined \def \showLCCN      #1{\unskip}     \fi
\ifx \shownote     \undefined \def \shownote      #1{#1}          \fi
\ifx \showarticletitle \undefined \def \showarticletitle #1{#1}   \fi
\ifx \showURL      \undefined \def \showURL       {\relax}        \fi
% invisible to TeX
\providecommand\bibfield[2]{#2}
\providecommand\bibinfo[2]{#2}
\providecommand\natexlab[1]{#1}
\providecommand\showeprint[2][]{arXiv:#2}

\bibitem[\protect\citeauthoryear{Bezerra, López-Ibáñez, and
  Stützle}{Bezerra et~al\mbox{.}}{2016}]%
        {bezerra2016automatic}
\bibfield{author}{\bibinfo{person}{Leonardo C.~T. Bezerra},
  \bibinfo{person}{Manuel López-Ibáñez}, {and} \bibinfo{person}{Thomas
  Stützle}.} \bibinfo{year}{2016}\natexlab{}.
\newblock \showarticletitle{Automatic Component-Wise Design of Multiobjective
  Evolutionary Algorithms}.
\newblock \bibinfo{journal}{\emph{IEEE Transactions on Evolutionary
  Computation}} \bibinfo{volume}{20}, \bibinfo{number}{3}
  (\bibinfo{year}{2016}), \bibinfo{pages}{403--417}.
\newblock
\urldef\tempurl%
\url{https://doi.org/10.1109/TEVC.2015.2474158}
\showDOI{\tempurl}


\bibitem[\protect\citeauthoryear{Blackwell and Branke}{Blackwell and
  Branke}{2006}]%
        {mQSO-2006}
\bibfield{author}{\bibinfo{person}{T. Blackwell} {and} \bibinfo{person}{J.
  Branke}.} \bibinfo{year}{2006}\natexlab{}.
\newblock \showarticletitle{Multiswarms, exclusion, and anti-convergence in
  dynamic environments}.
\newblock \bibinfo{journal}{\emph{IEEE Transactions on Evolutionary
  Computation}} \bibinfo{volume}{10}, \bibinfo{number}{4}
  (\bibinfo{year}{2006}), \bibinfo{pages}{459--472}.
\newblock
\urldef\tempurl%
\url{https://doi.org/10.1109/TEVC.2005.857074}
\showDOI{\tempurl}


\bibitem[\protect\citeauthoryear{Branke}{Branke}{1999}]%
        {MPB-1999}
\bibfield{author}{\bibinfo{person}{J. Branke}.}
  \bibinfo{year}{1999}\natexlab{}.
\newblock \showarticletitle{Memory enhanced evolutionary algorithms for
  changing optimization problems}. In \bibinfo{booktitle}{\emph{Proceedings of
  the 1999 Congress on Evolutionary Computation-CEC99 (Cat. No. 99TH8406)}},
  Vol.~\bibinfo{volume}{3}. \bibinfo{pages}{1875--1882 Vol. 3}.
\newblock
\urldef\tempurl%
\url{https://doi.org/10.1109/CEC.1999.785502}
\showDOI{\tempurl}


\bibitem[\protect\citeauthoryear{Branke}{Branke}{2002}]%
        {offline-error-2002}
\bibfield{author}{\bibinfo{person}{J{\"u}rgen Branke}.}
  \bibinfo{year}{2002}\natexlab{}.
\newblock \bibinfo{booktitle}{\emph{Empirical Evaluation}}.
\newblock \bibinfo{publisher}{Springer US}, \bibinfo{address}{Boston, MA},
  \bibinfo{pages}{67--98}.
\newblock
\showISBNx{978-1-4615-0911-0}
\urldef\tempurl%
\url{https://doi.org/10.1007/978-1-4615-0911-0_5}
\showDOI{\tempurl}


\bibitem[\protect\citeauthoryear{Campelo, Batista, and Aranha}{Campelo
  et~al\mbox{.}}{2020}]%
        {moeadr_paper}
\bibfield{author}{\bibinfo{person}{Felipe Campelo}, \bibinfo{person}{Lucas
  Batista}, {and} \bibinfo{person}{Claus Aranha}.}
  \bibinfo{year}{2020}\natexlab{}.
\newblock \showarticletitle{The {MOEADr} Package: A Component-Based Framework
  for Multiobjective Evolutionary Algorithms Based on Decomposition}.
\newblock \bibinfo{journal}{\emph{Journal of Statistical Software}}
  (\bibinfo{year}{2020}).
\newblock
\newblock
\shownote{In press. Available from: \url{https://arxiv.org/abs/1807.06731}}.


\bibitem[\protect\citeauthoryear{Clerc and Kennedy}{Clerc and Kennedy}{2002}]%
        {pso-parameters-2}
\bibfield{author}{\bibinfo{person}{M. Clerc} {and} \bibinfo{person}{J.
  Kennedy}.} \bibinfo{year}{2002}\natexlab{}.
\newblock \showarticletitle{The particle swarm - explosion, stability, and
  convergence in a multidimensional complex space}.
\newblock \bibinfo{journal}{\emph{IEEE Transactions on Evolutionary
  Computation}} \bibinfo{volume}{6}, \bibinfo{number}{1}
  (\bibinfo{year}{2002}), \bibinfo{pages}{58--73}.
\newblock
\urldef\tempurl%
\url{https://doi.org/10.1109/4235.985692}
\showDOI{\tempurl}


\bibitem[\protect\citeauthoryear{Eberhart and Shi}{Eberhart and Shi}{2000}]%
        {pso-parameters-1}
\bibfield{author}{\bibinfo{person}{R.C. Eberhart} {and} \bibinfo{person}{Y.
  Shi}.} \bibinfo{year}{2000}\natexlab{}.
\newblock \showarticletitle{Comparing inertia weights and constriction factors
  in particle swarm optimization}. In \bibinfo{booktitle}{\emph{Proceedings of
  the 2000 Congress on Evolutionary Computation. CEC00 (Cat. No.00TH8512)}},
  Vol.~\bibinfo{volume}{1}. \bibinfo{pages}{84--88 vol.1}.
\newblock
\urldef\tempurl%
\url{https://doi.org/10.1109/CEC.2000.870279}
\showDOI{\tempurl}


\bibitem[\protect\citeauthoryear{Fortin, {De Rainville}, Gardner, Parizeau, and
  Gagn\'e}{Fortin et~al\mbox{.}}{2012}]%
        {DEAP_JMLR2012}
\bibfield{author}{\bibinfo{person}{F\'elix-Antoine Fortin},
  \bibinfo{person}{Fran\c{c}ois-Michel {De Rainville}},
  \bibinfo{person}{Marc-Andr\'e Gardner}, \bibinfo{person}{Marc Parizeau},
  {and} \bibinfo{person}{Christian Gagn\'e}.} \bibinfo{year}{2012}\natexlab{}.
\newblock \showarticletitle{{DEAP}: Evolutionary Algorithms Made Easy}.
\newblock \bibinfo{journal}{\emph{Journal of Machine Learning Research}}
  \bibinfo{volume}{13} (\bibinfo{date}{jul} \bibinfo{year}{2012}),
  \bibinfo{pages}{2171--2175}.
\newblock


\bibitem[\protect\citeauthoryear{Hu and Eberhart}{Hu and Eberhart}{2002}]%
        {RPSO}
\bibfield{author}{\bibinfo{person}{Xiaohui Hu} {and} \bibinfo{person}{R.C.
  Eberhart}.} \bibinfo{year}{2002}\natexlab{}.
\newblock \showarticletitle{Adaptive particle swarm optimization: detection and
  response to dynamic systems}. In \bibinfo{booktitle}{\emph{Proceedings of the
  2002 Congress on Evolutionary Computation. CEC'02 (Cat. No.02TH8600)}},
  Vol.~\bibinfo{volume}{2}. \bibinfo{pages}{1666--1670 vol.2}.
\newblock
\urldef\tempurl%
\url{https://doi.org/10.1109/CEC.2002.1004492}
\showDOI{\tempurl}


\bibitem[\protect\citeauthoryear{Lavinas, Ladeira, Ochoa, and Aranha}{Lavinas
  et~al\mbox{.}}{2022}]%
        {lavinas_gecco2022}
\bibfield{author}{\bibinfo{person}{Yuri Lavinas}, \bibinfo{person}{Marcelo
  Ladeira}, \bibinfo{person}{Gabriela Ochoa}, {and} \bibinfo{person}{Claus
  Aranha}.} \bibinfo{year}{2022}\natexlab{}.
\newblock \showarticletitle{Component-Wise Analysis of Automatically Designed
  Multiobjective Algorithms on Constrained Problems}. In
  \bibinfo{booktitle}{\emph{Proceedings of the Genetic and Evolutionary
  Computation Conference}} \emph{(\bibinfo{series}{GECCO '22})}.
  \bibinfo{address}{New York, NY, USA}, \bibinfo{pages}{538–546}.
\newblock
\showISBNx{9781450392372}
\urldef\tempurl%
\url{https://doi.org/10.1145/3512290.3528719}
\showDOI{\tempurl}


\bibitem[\protect\citeauthoryear{López-Ibáñez, Dubois-Lacoste, {Pérez
  Cáceres}, Birattari, and Stützle}{López-Ibáñez et~al\mbox{.}}{2016}]%
        {LOPEZIBANEZ201643}
\bibfield{author}{\bibinfo{person}{Manuel López-Ibáñez},
  \bibinfo{person}{Jérémie Dubois-Lacoste}, \bibinfo{person}{Leslie {Pérez
  Cáceres}}, \bibinfo{person}{Mauro Birattari}, {and} \bibinfo{person}{Thomas
  Stützle}.} \bibinfo{year}{2016}\natexlab{}.
\newblock \showarticletitle{The irace package: Iterated racing for automatic
  algorithm configuration}.
\newblock \bibinfo{journal}{\emph{Operations Research Perspectives}}
  \bibinfo{volume}{3} (\bibinfo{year}{2016}), \bibinfo{pages}{43--58}.
\newblock
\showISSN{2214-7160}
\urldef\tempurl%
\url{https://doi.org/10.1016/j.orp.2016.09.002}
\showDOI{\tempurl}


\bibitem[\protect\citeauthoryear{Nguyen}{Nguyen}{2011}]%
        {survey-phd-2011}
\bibfield{author}{\bibinfo{person}{Trung Nguyen}.}
  \bibinfo{year}{2011}\natexlab{}.
\newblock \emph{\bibinfo{title}{Continuous dynamic optimisation using
  evolutionary algorithms}}.
\newblock \bibinfo{thesistype}{Ph.\,D. Dissertation}.
\newblock


\bibitem[\protect\citeauthoryear{Rohlfshagen and Yao}{Rohlfshagen and
  Yao}{2013}]%
        {related1-2013}
\bibfield{author}{\bibinfo{person}{Philipp Rohlfshagen} {and}
  \bibinfo{person}{Xin Yao}.} \bibinfo{year}{2013}\natexlab{}.
\newblock \showarticletitle{Evolutionary Dynamic Optimization: Challenges and
  Perspectives}. In \bibinfo{booktitle}{\emph{Evolutionary Computation for
  Dynamic Optimization Problems}},
  \bibfield{editor}{\bibinfo{person}{Shengxiang Yang} {and}
  \bibinfo{person}{Xin Yao}} (Eds.). \bibinfo{publisher}{Springer Berlin
  Heidelberg}, \bibinfo{address}{Berlin, Heidelberg}, \bibinfo{pages}{65--84}.
\newblock
\showISBNx{978-3-642-38416-5}


\bibitem[\protect\citeauthoryear{Topcuoglu, Ucar, and Altin}{Topcuoglu
  et~al\mbox{.}}{2014}]%
        {example2-2014}
\bibfield{author}{\bibinfo{person}{Haluk~Rahmi Topcuoglu},
  \bibinfo{person}{Abdulvahid Ucar}, {and} \bibinfo{person}{Lokman Altin}.}
  \bibinfo{year}{2014}\natexlab{}.
\newblock \showarticletitle{A hyper-heuristic based framework for dynamic
  optimization problems}.
\newblock \bibinfo{journal}{\emph{Applied Soft Computing}}
  \bibinfo{volume}{19} (\bibinfo{year}{2014}), \bibinfo{pages}{236--251}.
\newblock
\showISSN{1568-4946}
\urldef\tempurl%
\url{https://doi.org/10.1016/j.asoc.2014.01.037}
\showDOI{\tempurl}


\bibitem[\protect\citeauthoryear{Woldesenbet and Yen}{Woldesenbet and
  Yen}{2009}]%
        {algo-realoc-2009}
\bibfield{author}{\bibinfo{person}{Yonas~G. Woldesenbet} {and}
  \bibinfo{person}{Gary~G. Yen}.} \bibinfo{year}{2009}\natexlab{}.
\newblock \showarticletitle{Dynamic Evolutionary Algorithm With Variable
  Relocation}.
\newblock \bibinfo{journal}{\emph{IEEE Transactions on Evolutionary
  Computation}} \bibinfo{volume}{13}, \bibinfo{number}{3}
  (\bibinfo{year}{2009}), \bibinfo{pages}{500--513}.
\newblock
\urldef\tempurl%
\url{https://doi.org/10.1109/TEVC.2008.2009031}
\showDOI{\tempurl}


\bibitem[\protect\citeauthoryear{Yang and Li}{Yang and Li}{2010}]%
        {CPSO-2010}
\bibfield{author}{\bibinfo{person}{Shengxiang Yang} {and}
  \bibinfo{person}{Changhe Li}.} \bibinfo{year}{2010}\natexlab{}.
\newblock \showarticletitle{A Clustering Particle Swarm Optimizer for Locating
  and Tracking Multiple Optima in Dynamic Environments}.
\newblock \bibinfo{journal}{\emph{IEEE Transactions on Evolutionary
  Computation}} \bibinfo{volume}{14}, \bibinfo{number}{6}
  (\bibinfo{year}{2010}), \bibinfo{pages}{959--974}.
\newblock
\urldef\tempurl%
\url{https://doi.org/10.1109/TEVC.2010.2046667}
\showDOI{\tempurl}


\bibitem[\protect\citeauthoryear{Yazdani, Cheng, Yazdani, Branke, Jin, and
  Yao}{Yazdani et~al\mbox{.}}{2021a}]%
        {Survey_dynamicEAs-A}
\bibfield{author}{\bibinfo{person}{Danial Yazdani}, \bibinfo{person}{Ran
  Cheng}, \bibinfo{person}{Donya Yazdani}, \bibinfo{person}{Jürgen Branke},
  \bibinfo{person}{Yaochu Jin}, {and} \bibinfo{person}{Xin Yao}.}
  \bibinfo{year}{2021}\natexlab{a}.
\newblock \showarticletitle{A Survey of Evolutionary Continuous Dynamic
  Optimization Over Two Decades—Part A}.
\newblock \bibinfo{journal}{\emph{IEEE Transactions on Evolutionary
  Computation}} \bibinfo{volume}{25}, \bibinfo{number}{4}
  (\bibinfo{year}{2021}), \bibinfo{pages}{609--629}.
\newblock
\urldef\tempurl%
\url{https://doi.org/10.1109/TEVC.2021.3060014}
\showDOI{\tempurl}


\bibitem[\protect\citeauthoryear{Yazdani, Cheng, Yazdani, Branke, Jin, and
  Yao}{Yazdani et~al\mbox{.}}{2021b}]%
        {Survey_dynamicEAs-B}
\bibfield{author}{\bibinfo{person}{Danial Yazdani}, \bibinfo{person}{Ran
  Cheng}, \bibinfo{person}{Donya Yazdani}, \bibinfo{person}{Jürgen Branke},
  \bibinfo{person}{Yaochu Jin}, {and} \bibinfo{person}{Xin Yao}.}
  \bibinfo{year}{2021}\natexlab{b}.
\newblock \showarticletitle{A Survey of Evolutionary Continuous Dynamic
  Optimization Over Two Decades—Part B}.
\newblock \bibinfo{journal}{\emph{IEEE Transactions on Evolutionary
  Computation}} \bibinfo{volume}{25}, \bibinfo{number}{4}
  (\bibinfo{year}{2021}), \bibinfo{pages}{630--650}.
\newblock
\urldef\tempurl%
\url{https://doi.org/10.1109/TEVC.2021.3060012}
\showDOI{\tempurl}


\bibitem[\protect\citeauthoryear{Yazdani, Nasiri, Sepas-Moghaddam, and
  Meybodi}{Yazdani et~al\mbox{.}}{2013}]%
        {FTMPSO-2013}
\bibfield{author}{\bibinfo{person}{Danial Yazdani}, \bibinfo{person}{Babak
  Nasiri}, \bibinfo{person}{Alireza Sepas-Moghaddam}, {and}
  \bibinfo{person}{Mohammad~Reza Meybodi}.} \bibinfo{year}{2013}\natexlab{}.
\newblock \showarticletitle{A novel multi-swarm algorithm for optimization in
  dynamic environments based on particle swarm optimization}.
\newblock \bibinfo{journal}{\emph{Applied Soft Computing}}
  \bibinfo{volume}{13}, \bibinfo{number}{4} (\bibinfo{year}{2013}),
  \bibinfo{pages}{2144--2158}.
\newblock
\showISSN{1568-4946}
\urldef\tempurl%
\url{https://doi.org/10.1016/j.asoc.2012.12.020}
\showDOI{\tempurl}


\bibitem[\protect\citeauthoryear{Yazdani, Omidvar, Branke, Nguyen, and
  Yao}{Yazdani et~al\mbox{.}}{2019}]%
        {example3-2019}
\bibfield{author}{\bibinfo{person}{Danial Yazdani},
  \bibinfo{person}{Mohammmad~Nabi Omidvar}, \bibinfo{person}{Jurgen Branke},
  \bibinfo{person}{Trung Nguyen}, {and} \bibinfo{person}{Xin Yao}.}
  \bibinfo{year}{2019}\natexlab{}.
\newblock \showarticletitle{Scaling Up Dynamic Optimization Problems: A
  Divide-and-Conquer Approach}.
\newblock \bibinfo{journal}{\emph{IEEE Transactions on Evolutionary
  Computation}}  \bibinfo{volume}{PP} (\bibinfo{date}{03}
  \bibinfo{year}{2019}), \bibinfo{pages}{1--1}.
\newblock
\urldef\tempurl%
\url{https://doi.org/10.1109/TEVC.2019.2902626}
\showDOI{\tempurl}


\bibitem[\protect\citeauthoryear{Yazdani, Omidvar, Cheng, Branke, Nguyen, and
  Yao}{Yazdani et~al\mbox{.}}{2022}]%
        {GMPB-2022}
\bibfield{author}{\bibinfo{person}{Danial Yazdani},
  \bibinfo{person}{Mohammad~Nabi Omidvar}, \bibinfo{person}{Ran Cheng},
  \bibinfo{person}{Jürgen Branke}, \bibinfo{person}{Trung~Thanh Nguyen}, {and}
  \bibinfo{person}{Xin Yao}.} \bibinfo{year}{2022}\natexlab{}.
\newblock \showarticletitle{Benchmarking Continuous Dynamic Optimization:
  Survey and Generalized Test Suite}.
\newblock \bibinfo{journal}{\emph{IEEE Transactions on Cybernetics}}
  \bibinfo{volume}{52}, \bibinfo{number}{5} (\bibinfo{year}{2022}),
  \bibinfo{pages}{3380--3393}.
\newblock
\urldef\tempurl%
\url{https://doi.org/10.1109/TCYB.2020.3011828}
\showDOI{\tempurl}


\end{thebibliography}

%%
% %% If your work has an appendix, this is the place to put it.
%\newpage
%\clearpage
%\makeatother
%\appendix

% \subsection{Optimizer}

% \input{files/components/pso.tex}
% \input{files/components/es.tex}

% \subsection{Change detection}
% \input{files/components/reevaluation.tex}

% \subsection{Diversity control}
% \input{files/components/exclusion.tex}
% \input{files/components/local-search.tex}
% \input{files/components/anti-convergence.tex}

% \subsection{Population division and management}
% \input{files/components/multipopulation.tex}


% We implemented the operators and made them available at~\href{github rep}{github rep}.

\end{document}